# An Ontology-Based Dialogue Management System for Banking and Finance Dialogue Systems

**Duygu Altinok**
4Com Innovation Center
Berlin, Germany
duygu.altinok@4Com.de

**Abstract**

Keeping the dialogue state in dialogue systems is a notoriously difficult task. We introduce an ontology-based dialogue manager (OntoDM), a dialogue manager that keeps the state of the conversation, provides a basis for anaphora resolution and drives the conversation via domain ontologies. The banking and finance area promises great potential for disambiguating the context via a rich set of products and specificity of proper nouns, named entities and verbs. We used ontologies both as a knowledge base and a basis for the dialogue manager; the knowledge base component and dialogue manager components coalesce in a sense. Domain knowledge is used to track Entities of Interest, i.e. nodes (classes) of the ontology which happen to be products and services. In this way we also introduced conversation memory and attention *in a sense*. We finely blended linguistic methods, domain-driven keyword ranking and domain ontologies to create ways of domain-driven conversation. Proposed framework is used in our in-house German language banking and finance chatbots. General challenges of German language processing and finance-banking domain chatbot language models and lexicons are also introduced. This work is still in progress, hence no success metrics have been introduced yet.

## 1. Challenges with German Language

German language processing is inherently challenging in general, independent of what the specific NLP task is. The main challenge is high variability in word forms due to inflections and compound words.

Finance domain lexicons include many compound words just like other technical domains in German. Here are some examples from our in-house banking and finance lexicon:

| | |
|---|---|
| Verfügungsberechtigung | power to draw from an account |
| Sparkonto | savings account |
| Girokonto | checking account |
| Steuernummer | tax ID |
| Zahlungsverkehr | payment transactions |
| Zahlungsverkehrsraum | payment transactions area |
| Onlinebanking | online banking |
| Hypothekendarlehen | mortgage |

Nouns, adjectives and verbs can be inflected according to gender, number and person. Rich word forms can pose a challenge language understanding components. In this paper, we focus on dialogue management. However, one should keep in mind that input to dialogue management components are provided by natural language understanding components.

Another practical issue in everyday written language is the *umlaut* (mutated vowels). Everyday informal written text includes umlauts replaced by their plain counterparts i.e. "Madchen, uber, schon" rather than "Mädchen, über, schön" etc. Especially in conversational interfaces, usage of umlauts reduce significantly due to English layout keyboards or just being lax about punctuation while typing quickly on a smartphone. In our opinion, umlaut-to-plain vowel replaced words are also a part of chatbot language models.

Morphologically rich languages have received considerable attention from many researchers. Many technical papers have been published to highlight the inherent technical difficulties in statistical methods e.g. MT, ASR-TTS, language models, text classification; practical solutions are offered in (Mikolov et al., 2016).

We overcome the challenges of rich German morphology using DEMorphy, an open source German morphological analyzer and recognizer[1]. Throughout our work, all lemmatizing and morphological analysis tasks are done by DEMorphy.

## 2. Introduction

Keeping dialogue state in conversational interfaces is a notoriously difficult task. Dialogue systems, also known as chatbots, virtual assistants and conversational interfaces are already used in a broad set of applications, from psychological support[2] to HR, customer care and entertainment.

Dialogue systems can be classified into goal-driven systems (e.g. flight booking, restaurant reservation) vs open-domain systems (e.g. psychological support, language learning and medical aid). As dialogue systems has gained attention, research interest in training natural conversation systems from large volumes of user data has grown. Goal-driven systems admit slot-filling and hand crafted rules, which is reliable but restrictive in the conversation (basically the user has to choose one of available options). Open domain conversational systems, based on generative probabilistic models attracted attention from many researchers, due these limits for goal-oriented systems (Serban et al., 2015; Li et al., 2017).

---

[1] https://github.com/DuyguA/DEMorphy
[2] https://www.wysa.io

One problem with conversation is maintaining the dialogue state. This comprises of what the user said and how the chatbot answered, *what we're talking about* and which pieces of information are relevant to generating the current answer. Kumar et al. (2017) introduced neural networks with memory and attention (DMN). Done up to here DMN includes episodic memory and an attention module plus to a recurrent encoder decoder. DMN first computes question representation. Then the question representation triggers the attention process. Finally the memory module can reason the answer from all relevant information.

However, purely statistical approach has some drawbacks:
- statistical frameworks need **huge** training sets. Especially frameworks with many statistical components such as DMN, have a great number of parameters and are vulnerable to sparseness problems.
- Anaphora resolution is implicit. The anaphora resolutions go into neural network as implicit parameters, there's no direct easy way to see how the resolutions worked. Answers come through at least two distinct statistical layers, one encoder and one decoder at least. Thus there is no easy way to understand why a specific answer is generated and how the anaphora resolution contributed to the generation.

This paper addresses the dialog management. We will describe domain-driven ways to
- keep the conversation memory, both the user and the bot side
- make the anaphora resolution
- generate knowledge-based answers
- possibly contribute to *what to say next*
- integrate linguistic features into the context

NLU and answer generation modules will not be considered in detail in this paper. The focus is on how ontologies can be used to generate natural conversations. However we will present the outputs and presentations for clarity. The goal is here to improve quality of conversations via domain knowledge. This work is still in progress, hence we were not able to include performance metrics yet, given the difficult nature of evaluation of dialogue systems in general.

## 3. Proposed Framework

### 3.1 The Domain

This paper describes the methodology that is used in our in-house banking and finance chatbots. We chose the banking and finance domain due to rich set of products and specificity of proper nouns, named entities and verbs; high potential for disambiguate the context and drive the conversation. Though development was made on banking and finance domain, framework is applicable to other highly specific domains such as medicine, law and online-shopping.

Banking and finance conversations, either seeking financial advice or inquiring banking products; aim to *get information* rather than to accomplish a goal. Hence throughout this work, conversations are not goal-oriented but rather domain-driven. Users usually do not aim to achieve a well-defined goal, i.e. book a table at a restaurant or book a flight. The financial chat is mostly about getting information about rates, prices, investment instruments and sometimes about picking a suitable option i.e. *purchase advice*. Purchase rarely happens immediately after the advice. The banking chat can include both asking for information about account types, credit card types, their yearly fees and rates i.e. the banking products; or making money transfer, asking for account balance, viewing account activity i.e. achieving a certain goal. We will address these issues in next chapters.

### 3.2 Ontology-based Dialogues

Ontology-based conversating is indeed a way of domain-driven conversation. We used ontologies in our work for two purposes:
- to store the knowledge
- to navigate through the domain

The knowledge base component takes part in many dialogue systems. After the NLU component turns queries into a logical form, next step is to interrogate the knowledge base for the answer generation. In banking and finance domain ontologies, one potentially needs to store
- a range of banking products: credit, credit card, debit card…
- attributes of these products: general conditions, rates and fees
- range of banking services: ATM, money transfer...
- attributes of the banking services: ATM points, branch addresses...

For instance, in following conversations (Fig.1 & 2) it is not possible to generate an answer *without* knowing the product. The knowledge base module provides information to the answer generation module.

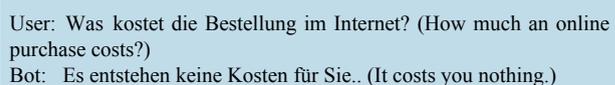

Figure 1: Dialogue Example

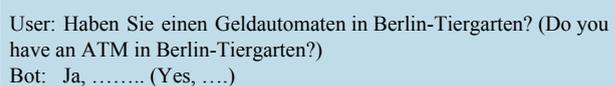

Figure 2: Dialogue Example

In our work the ontology also drives the conversation: *keeps the context*, provides a basis to the *anaphora resolution* and possibly produces *what to say next*. The knowledge base component and the dialogue management components coalesce.

> User: Ich möchte eine Kreditkarte bestellen. (I want to get a credit card.)
> Bot: Wir bieten Ihnen eine Mastercard an. (We have a Mastercard offer.)
> User: Was kostet die? (What does that cost?)
> Bot: …….
> User: Ist eine Internetbestellung möglich? (Is online purchase possible?)
> Bot: Ja, klicken Sie einfach auf "Mastercard beantragen" unter … (Yes, please click…)

Figure 3: Dialogue Example

In Figure 3, die refers to the Mastercard product. One needs to remember from the previous line that we were speaking about the Mastercard. The sentence "Was/PWS kostet/VVFIN die/PDS ?" is a wh-question sentence; also does not have any noun phrases, includes a substitution demonstrative pronoun and *nothing else*. Worse, the verb attaches to both the interrogative pronoun and the demonstrative pronoun. From the linguistic information so far, it is clear that one needs to resolve what die refers to. "*Ist eine Internetbestellung möglich?*" involves a more difficult kind of anaphora, an *implicit* one. There is a noun phrase "Internetbestellung", but it's a property (in semantic and ontological sense) of the Mastercard indeed; not any other banking product. If one *knows* that current *Entity of Interest*[3] is the Mastercard and the Mastercard has a property related to purchasing channels, then the resolution becomes clear.

We took FIBO as our guideline and created our own banking and finance domain ontologies in German. Including the whole ontology is not possible, thus here in Figure 4 we present a small section devoted to the credit products, generated by VOWL[4] within Protégé[5]:

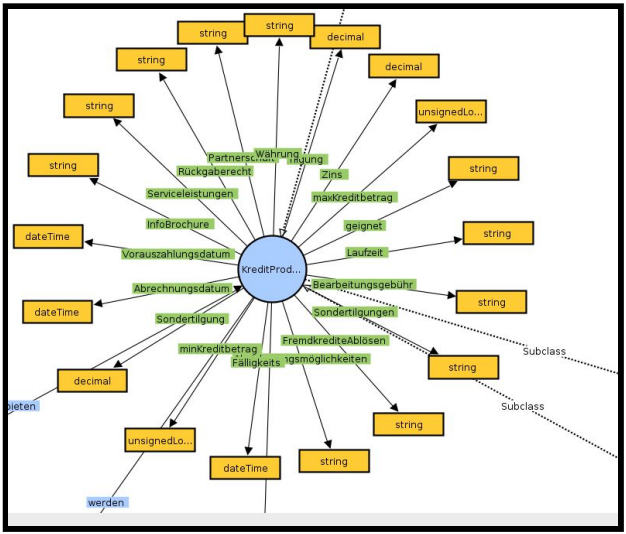

Figure 4: Ontology Example generated by VOWL within Protégé

---
[3] will be explained soon
[4] http://vowl.visualdataweb.org/protegevowl.html
[5] https://protege.stanford.edu/

Nodes i.e. the classes, consist of noun phrases (corresponding to abstract terms) and attributes (object and data properties) can be either noun phrases or verb phrases.

Domain ontology is an abstract schema. When we want to store information about a specific financial institute and deploy the corresponding chatbot, we instantiate the ontology with corresponding individuals. For instance, if we want to chat about 4Bank and its products, we instantiate the *KreditProduct* class by the *4Kredit* individual and the *Zins* with the *0.23* individual.

With the domain knowledge, we track significant entities in the conversation. The Entity of Interest notion includes *all classes* in the domain ontology (products and services). Noun phrases in the user inquiries fall into two categories, either a term from the ontology or not. Ontology terms are used to drive the conversation, the rest is kept as the user's conversation memory. This way we unite the linguistic knowledge, output of the NLU module with the domain knowledge. In every step of the conversation, the bot keeps track of user speaks about which product/service explicitly. Difference between the neural dialogue managers and the ontology-based approach becomes clear here, ontology-based DM keeps conversation memory **explicitly** throughout the conversation. In our framework, conversation memory and attention is provided via pointers to significant EOIs and keeping track of already visited nodes.

## 4. Implementation Details

Methods of resolution and disambiguation include blending linguistic information, several layers of semantic similarity and statistical methods together with ontologies. We will include the semantic representations of the user inputs to clarify the methodology. Throughout this paper, we will use 4Bank Ontology, our in-house banking and finance ontology.

### 4.1 The Domain Ontology

4Bank ontology contains 76 distinct classes, 37 object properties and 534 data properties. The ontology classes correspond to products, services, legal terms, contracts, customers as well as more abstract concepts such as "finance product". Abstract concepts usually serve as superclasses to more specific concepts. The object property (relation) names are verbs, the data property (attribute) names can be both names are verbs. The class names are always noun phrases; either proper nouns correspond to the products/services, or common nouns for more general concepts.

Reason for relatively less number of relations is that, many relations map superclasses of products/services to again superclasses. Figure 5 shows some examples of the classes from the ontology, together with their subclasses and identical classes:

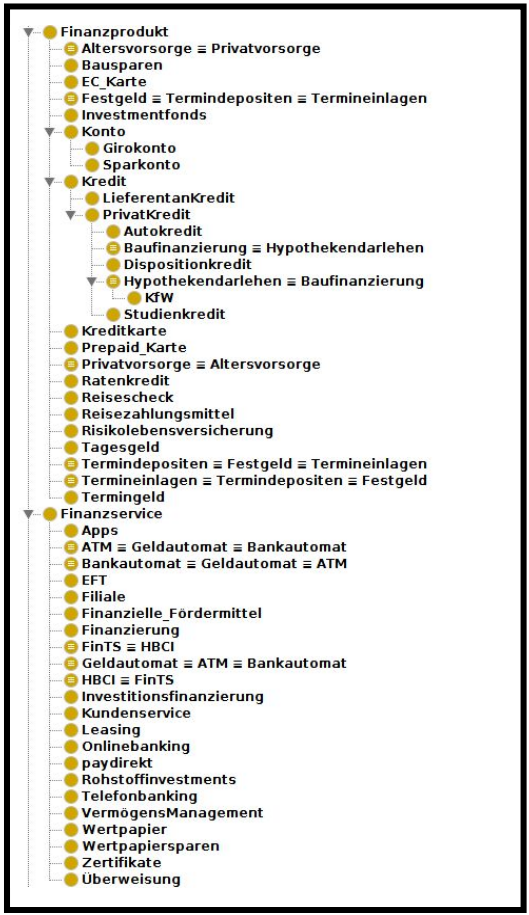

Figure 5: Classes of the Domain Ontology

Underlying undirected graph is dense and connected. Average degree of the underlying undirected graph is 3.1, maximum out- and in-degrees of underlying directed graph are 4 and 4, respectively (only the object properties are counted).

## 4.2 Architecture Overview

An overall architecture of our finance-banking FAQ chatbot is shown in Figure 6. First, the system transforms user's text message into a semantic representation. Then the semantic information is processed together with the current dialogue state to generate an appropriate answer.

We define a chat session as $S=(C, (Q_1, A_1, Q_2, A_2, ...))$; a finite sequence of user messages ($Q_i$) and bot answers ($A_i$), together with the current context ($C$).

During runtime, every new user message creates a new query object (throughout this paper we will use the user message and the corresponding query object interchangeably). The query object contains linguistic, semantic and syntactic information parsed from the user message. Only one context object is alive per chat session and the context object changes state with user queries.

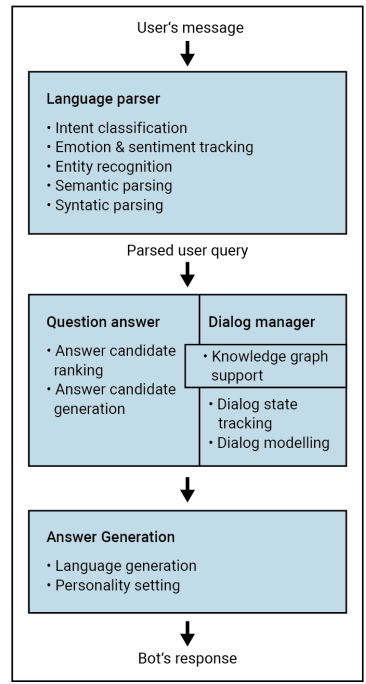

Figure 6: Architecture of the Banking FAQ Bot

### 4.2.1 Natural Language Understanding

The NLU component parses semantic and syntactic information from user queries. For the sake of simplicity, we will provide examples of single sentence user messages. Multi-sentence user messages are handled similarly. A query object has the following attributes: the sentence type, the intent, keyphrases, noun phrases, verb phrases, POS tags and the length as the number of tokens. There are four sentence types recognized by our system: Greeting, Chitchat, Action and Ordinary. Each type corresponds to a subclass of the query object. Some examples for each sentence types are given in Figure 7.

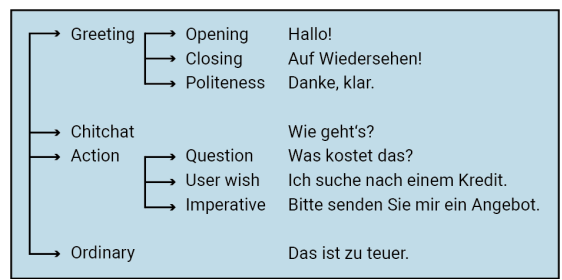

Figure 7: Sentence Types

Different subclasses might admit more fields. For instance, Action sentence has an "agent" and Question sentence has a "question type" (one of the following: yesno_q, wh_q or misc.)

Action sentences represent sentences that corresponds to an action, either from the user or from the bank. Action sentences also admits "agent"s. In our work, agent has a different meaning than its usual linguistic meaning. The agent is the performer of the action; either the user, the bank or a third person. See the examples:

```
User: Hallo!
{       'GTYPE': 'opening',
        'IS_UNINFORMATIVE': True,
        'LENGTH': 1,
        'SENTENCE': Hallo,
        'SENTENCE_TYPE': 'Greeting'}
```

Figure 8: Example Greeting Sentence

```
User: Können Sie mir Unterlagen zukommen lassen?
{
 'AGENT': '2nd',
 'INTENTS': ['send_document'],
 'IS_UNINFORMATIVE': False,,
 'KPHRASES': ['unterlagen', 'unterlagen zukommen lassen'],
 'LENGTH': 6,
 'NOUN_PHRASES': ['unterlagen'],
 'POS_TAGS':"VMFIN PPER PPER NN VVINF VVINF $.",
 'QTYPE': 'yesno_q',
 'QWORD': 'können',
 'SENTENCE': Können Sie mir Unterlagen zukommen lassen?,
 'SENTENCE_TYPE': 'Action',
 'TYPE_ATTR': 'question',
 'VERB_PHRASES': ['zukommen lassen']}
```

Figure 9: Example Question Sentence

Greeting and Chitchat sentences short-circuit to the answer generation module without further processing.

All syntactic parsing tasks were done by SpaCy. The semantic parsing comes in several layers, the question type, the keyphrases, the intent and the semantic similarity score based on word vectors during the graph search.

**Keyphrase Ranking** Keyphrase candidates are extracted in the NLU module by linguistic features. In our work we included noun phrases and noun phrases followed by verb phrases as keyphrase candidates. The IR module ranks candidates and generates the keyphrases. We designed an Okapi BM25 based keyword ranking algorithm in an unsupervised fashion. The ranking algorithm is trained on our in-house synonyms corpus. The corpus contains 120 questions in total, annotated with their rephrased (semantically similar) counterparts and keyphrases. See an example entry from the dataset:

Ich möchte meinen Kredit gerne erhöhen. Können Sie mir Unterlagen zukommen lassen?
Kredit erhöhen, Kredit aufstocken, Krediterhöhung, Kreditaufstockung, Unterlagen zur Kreditaufstockung, Unterlagen zur Krediterhöhung
Wie kann man den Kredit erhöhen
Wie kann ich meinen Kredit aufstocken
Wie kann ich den bestehenden Kredit aufstocken?
Ich möchte mein Darlehen erhöhen, Wie erhöhe ich mein Darlehen
Wie stockt man sein Darlehen auf
Senden Sie mir Unterlagen zur Krediterhöung
Schicken Sie mir Unterlagen zur Kreditaufstockung
Können Sie mir Formulare zur Krediterhöhung zukommen lassen

**Intent Classifier** The intent classifier is also trained on the synonyms dataset. The classifier is again subword based, we used fastText library[6] for the classification.

---

[6] https://pypi.python.org/pypi/fasttext

### 4.2.2 Context Object

The context object represents the current state of the conversation. The context object class has the following fields: curr_prod, curr_prod_indiv, curr_inode, curr_leaf and message_index.

The first four fields are pointers to the ontology nodes. The current product always refers to the product that has been the current chat topic and the corresponding curr_prod always points to one of subclasses of the Product or Service nodes. The current product individual references a specific instance of the current product, for instance if the current product class is the credit card, the current product individual can be Mastercard Gold or Mastercard Standard. The current inode points to the attribute of the current product class or individual spoken, the current leaf points to the most recently fetched leaf of the current inode if applicable. The current leaf can be both a class or an individual of the ontology. All four pointers reference significant EOIs of the chat; during the chat's lifetime we hold pointers to one individual, two conceptual classes as well as keeping track of the visited nodes.

A context object begins its lifetime with a null curr_prod and curr_prod_indiv. In every step of the conversation, if the current product or the current product instance is ambiguous or not present; the bot asks user for clarification or lists the available options to direct the conversation. See for instance:

```
U: Ich interesse für mich für ein Konto          curr_prod←Konto
B: Wir bieten Girokonten und Sparkonten an.
curr_prod has 2 subclasses Girokonto and Sparkonto. The bot lists all.
U: Was kostet das Girokonto?                    curr_prod←Girokonto
curr_prod is now Girokonto and has 2 individuals. The bot lists both.
B: Das Superkonto kostet 60 Euro pro Jahr und das Standard4Konto is
für Sie kostenlos.
```

Figure 10: Dialogue Example

This way we keep track of which product class, if applicable which specific product and which attributes are being discussed. It is also possible to conversate without referencing a specific individual, for instance:

```
U: Bieten Sie Leasing an? (Do you provide leasing?)
B: Nein, wir bieten nur Privatkredite an. (No, we provide only
personal kredits.)
```

Figure 11: Dialogue Example

### 4.3. Context Resolution Algorithm

In this, section we introduce our graph-based context resolution algorithm. The CRO algorithm is combination of depth-first search and breadth-first search on the ontology as a directed graph.

#### 4.3.1 Graph Search

We recall the example from Figure 12 in Section 3.2:

> User: Ich möchte eine Kreditkarte bestellen. (I want to get a credit card.)
> Bot: Wir bieten Ihnen eine Mastercard an. (We offer Mastercard.)

Figure 12: Dialogue Example

A short version of the output of NLU module for the first user input is below. "*Ich möchte eine Kreditkarte bestellen*" has only one noun phrase and "*Kreditkarte*" ranks as a keyphrase. Please see the output below:

> ActionS(type="user-wish",agent="customer",kphrases=["Kredit-karte", "Kreditkarte bestellen"], noun_phrases=["Kreditkarte"])

Figure 13: Output Example

Next step is to semantically match the noun phrases to the ontology. We search for the *Kreditkarte* in our 4Bank ontology and locate the class in the schema (see Figure 14):

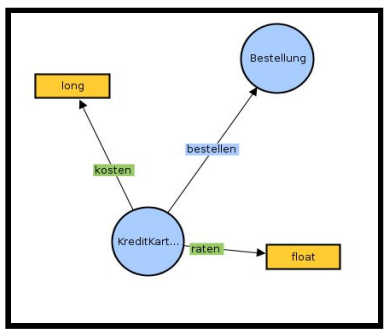

Figure 14: Matching Keyword *Kreditkarte* in the Ontology

Then *Kreditkarte* immediately becomes the curr_prod as it is a subclass of Finanzprodukt class. *Kreditkarte* has only one instance and it is called *Mastercard*. We feed this information to the answer generation module to generate an answer. Dialogue manager holds a pointer to this node to keep the context.

The graph search also includes word vector based semantic distance to match the synonyms correctly. For instance the keyphrase "Online" correctly matches to the node "Internet". We used fastText vectors[7] due to their superior performance on the German word forms.

### 4.3.2 Resolution

The context resolution algorithm resolves which specific product/service, concept or attribute the user is speaking about.

In every step of the context resolution algorithm,

(a) if a product individual name appears in the user query, we set curr_prod_indiv to this node and fetch the questioned attribute. We can directly fetch the answer from the ontology without any more resolution as the context is straightforward. Also the curr_prod is set the product individual's class.

---

[7] https://github.com/facebookresearch/fastText/blob/master/pretrained-vectors.md

> User: Was ist der Zinssatz bei 4Kredit? (What is the interest rate of 4Kredit?)     curr_prod←Kredit, curr_prod_indiv←4Kredit

Figure 15: Dialogue Example

(b) if the user mentions a product class, either it is the class name for a product individual that (s)he mentioned before or (s)he changes the context to speak more about this product class. In the latter case, we set curr_prod to the class name. In the first case, the class name refers to an already mentioned individual; hence curr_prod is already set to this class name.

> User: Hallo                                         curr_prod = null
> Bot: Hallo
> User: Ich bekomme einen Kredit.          curr_prod←Kredit
> Bot: Wir bieten Ihnen den 4Kredit an. (We provide 4Kredit.)

> User: Hallo                                         curr_prod = null
> Bot: Hallo
> User: Was ist der Zinssatz bei 4Kredit? (What 4Kredit's interest rate?)
> curr_prod←Kredit, curr_prod_indiv←4Kredit
> Bot: 0.23
> User: Was kostet der Kredit? (Here, Kredit refers to the individual 4Kredit. No updates to curr_prod or curr_prod_indiv are necessary)
> Bot: 4Kredit kostet ….

> User: Moin moin
> Bot: Hallo!
> User: Was ist der Zinssatz bei 4Kredit?
> curr_prod←Kredit, curr_prod_indiv←4Kredit
> Bot: 0.23
> User: Was kostet eine Kreditkarte? (What does a credit card cost?)
> (Here a context change occurs as the curr_prod_indiv 4Kredit is not an instance of Kreditkarte class.)      curr_prod←Kreditkarte

Figure 16: Dialogue Examples

(c) the user can query without specifying a product name or a product individual name. This usually happens at the beginning of the conversation:

> Bot: Hallo                                         curr_prod = null
> User: Hallo, was kostet das? (Hello, what does that cost?)
> curr_prod = null, curr_prod_indiv = null
> Bot: Wir bieten Kredite, Konten, … an. (We offer credits, accounts..)

Figure 17: Dialogue Example

In this case we list the available subclasses of the product/service classes and ask the user which one(s) (s)he is interested in.

(d) the user not always refers to a product/service explicitly. In this case the intent classifier comes into the play:

> User: Ich möchte meinen Hausbau finanzieren. Kannst du mir bitte helfen? (I want to finance my house construction. Would it be possible to help me?)        curr_prod←Hypothek

Figure 18: Dialogue Example

### 4.3.3 Pronominal Anaphora Resolution

We will continue with the example from section 4.3.1. We already make a graph search to locate the Kreditkarte node in the ontology and generated an answer to the first user question.

> User: Ich möchte eine Kreditkarte bestellen. (I want to get a credit card.)
> Bot: Wir bieten Ihnen eine Mastercard an. (We offer Mastercard.)
> User: Was kostet die? (What does that cost?)
> Bot: 80 Euro jährlich (80 euros annually.)

Figure 19: Dialogue Example

Next user input is "*Was/PWS kostet/VVFIN die/PDS ?*". Please see the semantic representation and the result of the dependency parse[8] in Figure 20 below:

> Q(type="wh-q", agent="no", qword="was")

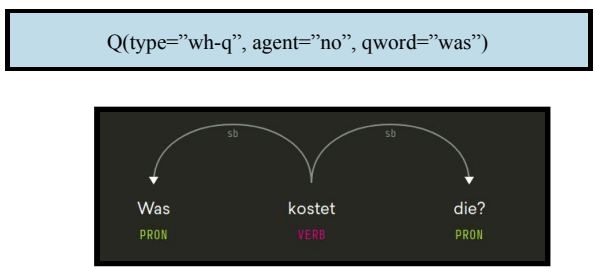

Figure 20: Example of the User Input

There are two pronouns in the sentence, an interrogative pronoun *Was*, one substitution demonstrative pronoun *die* and a verb. There are no noun phrases in the sentence. It becomes clear that this question does not carry much *information* itself and die needs to be resolved. In the current context, we have only one past EOI i.e. the "*Kreditkarte*". After lemmatizing with DEMorphy, "kostet" becomes "kosten" and resolves to the "kosten" relation in the domain ontology, please see Figure 20. The KB module fetches the individual related, "*80 Euro jährlich*" and inputs it to the rest of the pipeline.

### 4.3.4 Implicit Anaphoras

It is not always clear when one should do any anaphora resolution at all. Following conversation in Figure 19 includes two such user inputs, where necessity for any anaphora resolution is not immediately recognizable.

> User: Kann ich einen Kredit aufnehmen? (Can I get a loan?)
> Bot: Die 4Bank bietet Ihnen mit 4Kredit einen schnellen und unkomplizierten Weg zu Ihrem Kredit. (4Bank offers 4Kredit, a fast and uncomplicated way to obtain a credit.)
> User: Ist eine Internetbestellung möglich? (Is online purchase possible?)
> Bot: Ja, klicken Sie …. (Yes, please click…)
> User: Am Telefon? (Purchase by phone?)
> Bot: Rufen Sie uns unter 05113003990 an und vereinbaren Sie einen Termin für Ihre telefonische Kreditbestellung. (Please phone us....)

Figure 21: Dialogue Example with Ambiguous Anaphora

---
[8] generated by spaCy and displaCy, https://spacy.io/

Question in Line 3 contains *information*, indeed EOI that appears in our banking and finance lexicon:

> Q(type="yesno-q",agent="no",qword="ist",kphrases=["Internet-bestellung"])

Figure 22: Semantic Parse of the User Input

Obviously this question is about purchasing online but it is still not clear only from this line, user is interested in purchasing *which* product online. Should it be a credit or an insurance? If there are several products that can be purchased online, it is not possible to generate an answer only based on this question. Here the context memory i.e. the current product ("*Kredit*") comes into play. The context resolution algorithm first performs a BFS to locate Bestellung node. Head of a German compound is its last word, previous words might qualify the head word in different ways. Internetbestellung has Bestellung as the head word. The current inode pointer is set to the Bestellung node and another BFS is performed to locate Internet subclass to mark Internet as the current leaf. Then the answer generator can take into account that (1) it is a yes-no question and (2) Bestellung class has a subclass called Internet to generate the answer "yes".

At first glance it is not clear that one needs a context disambiguation at all. However, these type of ambiguities might appear even at the very final stages of the context disambiguation computations.

It is also possible that there are more than one EOIs in the current context memory. Line 5 is an example for such situation, current EOIs are the *Kredit* and the Internetbestellung. Telefon indeed is a "shorthand" for the *Telefonbestellung* (purchase by phone). Here Telefon semantically attaches to the both EOIs, the *Kredit* (as an attribute) and the Internetbestellung (not directly, via analogy) (see Fig. 22).

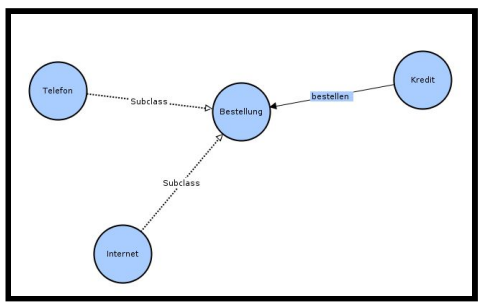

Figure 23: Example of EOI Ambiguity

The CRO handles this case as follows: The current leaf is the Internet, the current inode is the Bestellung. Previous keyword which resolved to Internet node was Internetbestellung, the concatenation of the node itself (Internet) and its parent node (Bestellung). Then, the context resolution algorithm first looks for a Telefon subclass under the Bestellung node. If it was the case no Telefon node was located under the current inode, the CRO would perform another BFS under the current product node, the *Kredit*. The resolution algorithm favors more specific anaphoras and the most recent EOI.

This way the CRO resolves the implicit semantic analogy that Telefon in Line 5 refers to Telefonbestellung and finally fetch the correct node.

We already pointed out that our framework introduces memory and attention *in a sense*. Dynamic memory networks with attention *selects* relevant previous episodic memory members, the *relevancy* is computed by the attention module. Our framework offers a more explicit solution than the statistical methodology, since at every step of the conversation the DM stores EOIs explicitly.

### 4.3.5 Long Sentences with Compound Intents

Multi-sentence user messages and long sentences may contain more than one intents and EOIs. For instance:

> Sind meine Unterlagen schon bei Ihnen eingegangen? Falls ja, wo erfahre ich den aktuellen Bearbeitungsstand? (Have you already received my documents? If so, where do I get the current processing status?)

> Wo und wie bekomme ich eine Kredit? (Where and how can I get a credit?)

> Was kostet eine Kredit und zu welcher Laufzeit? (What does a credit cost and what is the loan period?)

Figure 24: Dialogue Examples

In the first sentence, there is one EOI, the *Unterlagen* and two intents: to check out if the documents arrived and to learn the processing status. The current product or the current product individual does not need any updates. The CRO sets the curr_inode to Unterlagen and answer generation module generates two answers for the two different intents and concatenate them to generate a composite answer.

The second sentence involves a product class name, *Kredit*. The CRO sets the current product to *Kredit* and the answer generation module generates an answer for the two intents jointly unlike the first sentence.

A product class name (*Kredit*) and two different attribute names of this class (*kosten*, *Laufzeit*) comes together in the third sentence. The context resolution rule (b) from the Section 4.3.2 applies here, we fetch two attributes the *kosten* and the *Laufzeit* iteratively and pass to the answer generation module.

However, even the sentence is long, it is very unusual to refer to two or more product classes and instances. For instance, such a sentence

> Was kostet ein Kredit und eine Kreditkarte? (What is the cost of a credit and a credit card?)

Figure 25: Dialogue Example

with two distinct product class names occurred almost none times in our evaluation corpus.

### 4.3.6 Cycle Prevention

While traversing the graph, the context resolution algorithm marks the visited nodes to prevent the future cycles. Users may ask the same questions that they previously asked, however the bot shouldn't make the same offerings that it already made. For instance:

> Bot: Hallo!
> User: Was bieten Sie an? (What services do you offer?)
> Bot: Wir bieten Kredite, Konto und Kreditkarte für Sie an.
> User: Was kostet der Kredit? (How much does the credit cost?)
> (Here the CRO marks Kredit node as visited.)
> Bot: …….
> User: …..
> Bot: ….
> User: Welche anderen Produkte haben Sie? (What else do you offer?)
> (Kredit node is marked as visited. The BFS skips this node.)
> Bot: Möchten Sie unseren Kreditkarten oder Kontobereich besuchen?
> User: ….

Figure 26: Dialogue Example

The CRO also marks the edges (data properties) that have already lead to fetching an answer:

> B: Hallo
> U: Ich brauche einen Kredit. (I need a credit.)
> B: Wir bieten Ihnen 4Kredit an. (We offer 4Kredit.)
> U: Was ist die Laufzeit? (What is the loan period?)
> (Here Laufzeit edge is marked as used)
> B: Wir bieten Laufzeiten von 12 bis 84 Monaten an. Wie hoch ist Dein Kreditwunsch? (The loan period is between 12 and 84 months. What amount are you asking for?)
> (Since we already fetched the Laufzeit edge, we do not speak about this attribute anymore. The BFS skips Laufzeit edge and the DM generates *what to say next* from another attribute: the kredit amount ).

Figure 27: Dialogue Example

The current product node and previous curr_prod nodes are always marked as visited. Marked product nodes and attributes together prevents back edges, hence cycles in the chat session.

## 5. Conclusion and Future Work

This paper presents methods of ontology-based ways of dialogue management in the banking and finance area. Though this work has not been tested extensively yet, the current achievements are promising. The future work includes finishing the framework and the chatbot development. Besides, it is even more important that we will evaluate the whole chatbot success by the BLEU score. We also aim to introduce a metric for the evaluation of the dialogue manager module. Considering the fact that the quality of the ontology directly affects the success of the DM, introducing an evaluation metric is a separate application on its own.

# 6. Bibliographical References